\documentclass[lettersize,journal]{IEEEtran}
\usepackage[colorlinks,urlcolor=blue,linkcolor=blue,citecolor=blue]{hyperref}
\usepackage{comment}
\usepackage{color,array}
\usepackage[utf8]{inputenc} 
\usepackage[T1]{fontenc}    
\usepackage{hyperref}       
\usepackage{url}            
\usepackage{booktabs}       
\usepackage{amsfonts}       
\usepackage{nicefrac}       
\usepackage{microtype}      
\usepackage{xcolor}         
\usepackage{amsmath}        
\usepackage[square,numbers]{natbib}
\usepackage{hyperref}
\usepackage{lineno}
\usepackage{amssymb}
\usepackage{upgreek}
\usepackage{multirow}
\usepackage{amsmath} 
\usepackage{xcolor}
\usepackage{algorithm}
\usepackage{algpseudocode}
\usepackage{bm}
\usepackage{calrsfs}
\usepackage{autobreak}
\usepackage{dirtytalk}
\usepackage[normalem]{ulem}
\usepackage{xr}
\usepackage{graphicx}
\usepackage{xcolor,colortbl}
\usepackage{caption}
\usepackage{tabu}
\usepackage{subcaption}
\usepackage{tablefootnote}
\usepackage{footnote}
\usepackage[symbol]{footmisc}

\newtheorem{definition}{Definition}[section]

\DeclareMathAlphabet{\pazocal}{OMS}{zplm}{m}{n}
\newcommand{\Manifold}{\pazocal{M}}

\newcommand{\Tau}{\pazocal{T}}
\newcommand{\Dataset}{\pazocal{D}}

\newcommand{\gfun}{\pazocal{G}}
\newcommand{\Loss}{\pazocal{L}}
\newcommand{\RNum}[1]{\uppercase\expandafter{\romannumeral #1\relax}}
\DeclareMathOperator{\vect}{vec}

\definecolor{electricgreen}{rgb}{0.0, 1.0, 0.0}
\definecolor{auburn}{rgb}{0.43, 0.21, 0.1}
\definecolor{darkgreen}{rgb}{0.0, 0.2, 0.13}
\definecolor{blue(ryb)}{rgb}{0.0, 0.0, 1.0}
\definecolor{greybck}{rgb}{0.3, 0.3, 0.3} 
\definecolor{greybck}{rgb}{0.5, 0.5, 0.5}
\definecolor{carnationpink}{rgb}{1.0, 0.65, 0.79}
\def\BibTeX{{\rm B\kern-.05em{\sc i\kern-.025em b}\kern-.08em
    T\kern-.1667em\lower.7ex\hbox{E}\kern-.125emX}}
\usepackage{balance}
\begin{document}

\title{FORML: A Riemannian Hessian-free Method for Meta-learning on Stiefel Manifolds}

\author{H. Tabealhojeh, S. Kumar Roy, P. Adibi and H. Karshenas
\thanks{H. Tabealhojeh is with the Department of Artificial Intelligence, Faculty of Computer Engineering, University of Isfahan, Isfahan, Iran (e-mail: h.tabealhojeh@eng.ui.ac.ir).}
\thanks{S. Kumar Roy is with Computer Vision Laboratory, EPFL, Switzerland (e-mail: soumava.roy@epfl.ch).}
\thanks{Corresponding author: P. Adibi, is with the Department of Artificial Intelligence, Faculty of Computer Engineering, University of Isfahan, Isfahan, Iran (e-mail: adibi@eng.ui.ac.ir).}
\thanks{H. Karshenas is with the Department of Artificial Intelligence, Faculty of Computer Engineering, University of Isfahan, Isfahan, Iran (e-mail: h.karshenas@eng.ui.ac.ir).}

\thanks{Manuscript received April 19, 2021; revised August 16, 2021.}}

\markboth{Journal of \LaTeX\ Class Files,~Vol.~14, No.~8, August~2021}%
{Shell \MakeLowercase{\textit{et al.}}: A Sample Article Using IEEEtran.cls for IEEE Journals}

\maketitle
\IEEEpeerreviewmaketitle
\begin{abstract}
  Meta-learning problem is usually formulated as a bi-level optimization in which the task-specific and the meta-parameters are updated in the inner and outer loops of optimization, respectively. However, performing the optimization in the Riemannian space, where the parameters and meta-parameters are located on Riemannian manifolds is computationally intensive. Unlike the Euclidean methods, the Riemannian backpropagation needs computing the second-order derivatives that include backward computations through the Riemannian operators such as retraction and orthogonal projection. This paper introduces a Hessian-free approach that uses a first-order approximation of derivatives on the Stiefel manifold. Our method significantly reduces the computational load and memory footprint. We show how using a Stiefel fully-connected layer that enforces orthogonality constraint on the parameters of the last classification layer as the head of the backbone network, strengthens the representation reuse of the gradient-based meta-learning methods. Our experimental results across various few-shot learning datasets, demonstrate the superiority of our proposed method compared to the state-of-the-art methods, especially MAML, its Euclidean counterpart.         
\end{abstract}

\begin{IEEEkeywords}
Few-shot classification, Riemannian optimization, Meta-learning, Stiefel manifold
\end{IEEEkeywords}
\section{Introduction}
\label{Sec-Introduction}
\IEEEPARstart{M}{eta-learning} has attracted growing attention recently, owing to its capability to learn from several tasks and generalize to unseen ones. From the optimization-based point of view, meta-learning can be formulated as a bi-level optimization problem, consisting of two optimization levels. At the outer level, a meta-learner extracts the meta-knowledge (the \textit{how to learn } knowledge) from a number of specific tasks, whereas at the inner level, each specific task is solved via a base-learner. Among different approaches to meta-learning \cite{hospedales2021meta}, in optimization-based methods the meta-learner \textit{learns how to optimize} the task-specific base learners. In other words, both the meta and base learners are usually gradient descent-based optimizers. Then, the meta-knowledge can be the step-size, gradient information, or anything that can affect or improve the inner-level optimizers. 

In practice, numerous learning tasks are modeled as optimization problems with nonlinear constraints. For instance, many classical machine learning problems such as Principal Component Analysis (PCA) \cite{yuan2019online}, Independent Component Analysis (ICA) \cite{hyvarinen2000independent} and subspace learning \cite{chakraborty2020intrinsic} are cast as optimization problems with orthogonality constraints. Also, modern deep neural networks, like convolutional neural networks (CNNs), benefit from applying the orthogonality constraint on their parameter matrices, which results in significant advantages in terms of accuracy and convergence rate \citep{bansal2018can}, generalization \citep{cogswell2015reducing} and distribution stability of the of neural activations during thier training \citep{huang2018building}. A popular way to incorporate nonlinear constraints into optimization frameworks is to make use of Riemannian geometry and formulate constrained problems as an unconstrained optimization on Riemannian manifolds \cite{lin2008riemannian}. In this case, Euclidean meta-learning with nonlinear constraints is cast as meta-learning on Riemannian manifolds. To do so, the optimization algorithm should be respectful to the Riemannian geometry, which can be achieved by utilizing the Riemannian operations such as \textit{retraction, orthogonal projection} and \textit{parallel transport} \cite{roy2018geometry}. For the orthogonality constraints, the search space of the problem can be formulated on the Stiefel manifold, where each point on the manifold is an orthonormal matrix. Hadi et. al introduced RMAML, a Riemannian optimization-based meta-learning method for few-shot classification based on this idea \cite{tabealhojeh2023rmaml}. Although RMAML is a general framework and can be applied to a wide range of Riemannian manifolds, we focused on meta-learning with orthogonality constraints, which is performed on the Stiefel manifold, demonstrating the effectiveness of enforcing the orthogonality constraints on the parameters and meta-parameters of the backbone network in terms of convergence rate and robustness against over-fitting. However, optimization-based meta-learning (even in Euclidean space) needs second-order differentiation that passes through the inner-level optimization path, which is complex and computationally expensive. The problem becomes more critical in the Riemannian spaces, where nonlinear operations such as retraction and orthogonal projection are used for the optimization. Moreover, each point on the Riemannian manifold is a matrix (opposite to Euclidean spaces where every point is a vector), which increases the computational load and memory footprint.    

In this paper, we propose a Riemannian bi-level optimization method that uses a first-order approximation to calculate the derivatives on the Stiefel manifold. In the inner loop of our algorithm, base learners are optimized using a limited number of steps on the Riemannian manifold, and at the outer loop, the meta-learner is updated using a first-order approximation of derivatives for optimization.  
In this way, our method avoids the exploding gradient issue caused by the Hessian matrices, resulting in the computational efficiency and stability of the bi-level optimization. We empirically show that our proposed method can be trained efficiently and learn a good optimization trajectory compared to the other related methods. We use the proposed method to meta-learn the parameters and meta-parameters of the head of the backbone model (the last fully-connected layer of the model) on the Stiefel manifold and demonstrate how normalizing the input of the head leads to computing the cosine similarities between the input and the output classes, which results in increasing the representation reuse phenomena, when the task-specific base-learners get optimized in the inner-loop utilizing the representation learned by the meta-learner.

The rest of the paper is structured as follows. The literature on optimization-based meta-learning, Riemannian meta-learning, and optimization techniques on Riemannian spaces are outlined in Section~\ref{Sec-related-works}. The foundation principles on which we will base our proposed algorithms are discussed in Section~\ref{Sec-Preliminaries}. Section~\ref{Sec-method} describes our proposed technique, named First Order Riemannian Meta-Learning (FORML) method. Section~\ref{Sec-Results} is dedicated to the presentation of experimental analysis, evaluations, and comparative results. Finally, Section~\ref{Sec-Conclusion}  provides a conclusion of the paper.
\section{Related Works}
\label{Sec-related-works}
\subsection{Optimization-based meta-learning}
Gradient-Based Meta-Learning (GBML) methods are a family of optimization-based meta-learning techniques in which both the meta and base learners are gradient-based optimizers. 
MAML \cite{finn2017model} is a highly popular optimization-based meta-learning method that has achieved competitive results on several few-shot learning benchmarks. MAML solves the bi-level optimization problem using two nested loops. At the inner loop, MAML learns the task-specific parameters, and the outer loop finds the universal meta-parameters that are used as initialization for the inner loop parameters. Using the meta-parameters as an initial point, the task-specific learning of the inner loop can be quickly done using only a few (gradient descent-based) optimization steps and a small number of samples. Based on several recent studies, the success of MAML can be explained via two concepts: feature reuse and feature adaptation. Feature reuse refers to the high-quality features generated by the meta-learners, that meta-initialize the task-specific features of the base learners. Based on the feature reuse idea, Raghu et. al. \cite{raghu2019rapid} have introduced ANIL, a variation of MAML, and claimed that MAML can be trained with almost no inner loop. Their method is similar to MAML, but they train only the head (last fully connected) layer of the base learners (that are initialized by the meta-parameters) and reuse the features of the meta learner. In other words, during the inner loop of ANIL, only the head of the base learner is updated and the other layers remain frozen (unchanged). However, Oh et. al. \cite{oh2021boil} have introduced BIOL method, and focused on causing a significant \textit{representation change} in the body of the model rather than \textit{representation reuse} in the head. Thus, during the inner loop, they have frozen the head of the base learners, enforcing the rest of the layers to change significantly.
\subsection{Riemannian meta-learning}
Recently, many researchers have focused on developing meta-learning methods on Riemannian manifolds \cite{tabealhojeh2023rmaml,gao2020learning,gao2022learning}. 
Tabealhojeh et. al. introduced RMAML, a framework for meta-learning in Riemannian spaces \cite{tabealhojeh2023rmaml}. Their method can be regarded as the Riemannian counterpart of MAML. They applied RMAML on the few-shot classification setting and showed the efficiency of meta-learning with orthogonality constraints (imposed by the Stiefel manifold). Gao et. al. \cite{gao2020learning} developed a learning-to-optimize method that acts on SPD manifolds. Their idea is based on using a recurrent network as a meta-learner. To do so, they introduced a \textit{matrix LSTM} to enable the recurrent network to act on the SPD manifold. For Riemannian manifolds other than SPD manifold and to reduce the computational complexity burden of their method, they introduced a generalization of their previous method, where an implicit method is used to differentiate the task-specific parameters with respect to the meta-parameters of the outer loop, avoiding differentiating through the entire inner-level optimization path \cite{gao2022learning}. However, except RMAML, other Riemannian methods have not focused on few-shot classification and only have experimented their methods on the classical optimization problems on Riemannian manifolds. In this work, similar to RMAML, we focus on few-shot classification and show how using Stiefel manifolds in meta-learning impacts solving these problems.  
\subsection{Hessian matrix, the curse of optimization-based methods}
Although optimization-based methods have achieved state-of-the-art performance in meta-learning, they suffer from the high computational cost needed for backpropagation. The bi-level optimization problem in these methods which is usually solved using two nested loops, requires gradients of the task-specific parameters in the inner loopto be propagated back for updating the meta-parameters of the outer loop. Computing these gradients needs differentiating through the chain of the entire inner loop updates, leading to Hessian-vector product. Furthermore, in Riemannian spaces, the nonlinear essence of Riemannian optimization will be more complex and computationally expensive, because the differentiation chain will pass through the complicated Riemannian operations such as retraction and orthogonal projection.          

Several papers in the literature have focused on reducing the computation complexity of optimization-based methods. Finn et al. \cite{finn2017model} introduced FOMAML, a first-order approximation for MAML, by ignoring the second-order Hessian matrix terms. The experimental results demonstrated the computational efficiency of FOMAML and showed that the few-shot accuracy is competitive with the original MAML. Reptile is another Hessian-free algorithm that updates the meta-parameters of the meta-learner (the outer loop) toward the task-specific parameters of base-learners \cite{nichol2018first}. Regarding meta-learning in Riemannian spaces, despite the good experimental results of the implicit differentiation approach proposed by Gao et al. \cite{gao2022learning}, using a matrix-LSTM as a meta-learner requires a huge amount of meta-parameters and heavy computational resources. Using the same meta-learning structure in terms of meta and base learners, Fan et al. \cite{fan2021learning} proposed to perform the inner-level optimization on the tangent space of the manifolds. As the tangent space of every point of a Riemannian manifold is itself an Euclidean hyper-plane, the inner-level optimization is performed on the Euclidean space, without the need to utilize the complex Riemannian operations. Their proposed method has not been experimented on few-shot classification benchmarks and is developed for learning to optimize classic problems in the Riemannian space such as the PCA optimization problems.

In this research, we introduce a first-order optimization-based meta-learning method formulated on Stiefel manifolds and study the effectiveness of orthogonality constraints enforced by the Stiefel manifold in few-shot classification. Moreover, we demonstrate the superiority of our proposed method against both \textit{representation reuse} and \textit{representation change} methods.   
\section{Preliminaries}
\label{Sec-Preliminaries}

\subsection{Riemannian manifolds}
In this section, we introduce the definition of smooth Riemannian manifolds and the required mathematical operations for applying Gradient Descent (GD) or GD-based optimization algorithms in Riemannian spaces.    
\begin{definition}[Smooth Riemannian manifold]
A smooth Riemannian manifold $\Manifold$ is a locally Euclidean space and can be understood as a generalization of the notion of a surface to higher dimensional spaces \cite{lin2008riemannian}. For each point $\textit{\textbf{P}}\in \Manifold$, the tangent space is denoted by $T_\textit{\textbf{P}}\Manifold$, which is a vector space that consists of all vectors that are tangent to $\Manifold$ at $\textit{\textbf{P}}$.
\end{definition}
\begin{definition}[Stiefel manifold]
A Stiefel manifold $St(n,p), n \geq p$, is a Riemannian manifold that is composed of all $n \times p$ orthonormal matrices, i.e. $\{\textbf{P}\in \mathbb{R}^{n \times p}:\textbf{P}^T\textbf{P}=\textbf{I}\}$. 
\end{definition}

\begin{definition}[Manifold optimization]
Manifold optimization, is an essential framework for tackling complex optimization challenges where the parameters exhibit nonlinearity and interdependencies that can be effectively represented and optimized within the smooth geometric structure of a Riemannian manifold. By leveraging the intrinsic properties of the manifold, such as its curvature and metric, manifold optimization techniques offer a powerful approach to navigating the search space of parameters and finding optimal solutions for a diverse range of real-world problems.
The nonlinear optimization problems can be formed as:
\begin{equation}
\label{eqn:mani_optimization}
\min_{\bm{P}\in\Manifold}{J(\bm{P})} 
\end{equation}
In this context, $J$ represents the loss function that needs to be minimized, which is parameterized by a set of constrained parameters denoted as $ \bm{P}$ exist on a smooth Riemannian manifold \( \Manifold \) \cite{absil2009optimization}. 
\end{definition}

\begin{figure*}[t]
\centering
\includegraphics[scale=0.13, keepaspectratio]{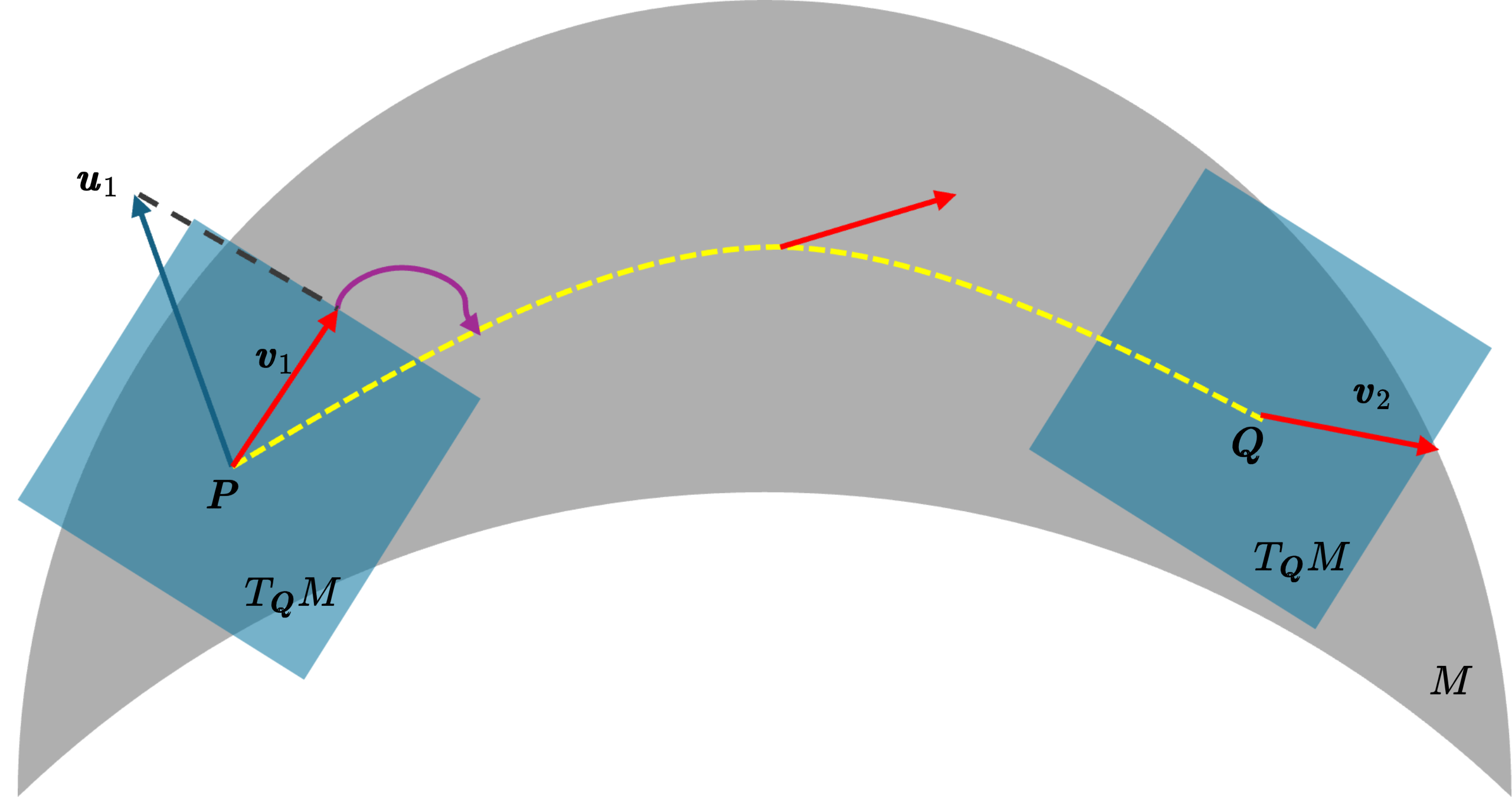}
\caption{An Explanatory diagram illustrating the Riemannian operations necessary for GD-based Riemannian optimization. Let $\textit{\textbf{P}}$ and $\textit{\textbf{Q}}$ denote points on the manifold $\Manifold$ linked by a geodesic represented by the yellow dashed curve. The tangent spaces at $\textit{\textbf{P}}$ and $\textit{\textbf{Q}}$ are $T_{\textit{\textbf{P}}} \Manifold$ and $T_{\textit{\textbf{Q}}} \Manifold$, respectively and are shown by blue color. Vector $\textit{\textbf{v}}_1 \in T_{\textit{\textbf{P}}} \Manifold$ is obtained through the orthogonal projection of the Euclidean vector $\textit{\textbf{u}}$ at point $\textit{\textbf{P}}$. The retraction $\textit{\textbf{R}}\!=\!R_{\textit{\textbf{P}}}(\textit{\textbf{v}}_1)$ is utilized for returning to the manifold from the tangent space at $\textit{\textbf{P}}$. Within a vicinity of $\textit{\textbf{P}}$, the retraction operation (depicted in purple) shows a location on the geodesic. The process of parallel transport $\textit{\textbf{v}}_2\!=\!\Gamma_{\textit{\textbf{P}}\rightarrow \textit{\textbf{Q}}}(\textit{\textbf{v}}_1)$ maps $\textit{\textbf{v}}_1 \in T_{\textit{\textbf{P}}} \Manifold$ to $\textit{\textbf{v}}_2 \in T_{\textit{\textbf{Q}}} \Manifold$ by moving in parallel along the geodesic (displayed with red arrows) connecting points \textit{\textbf{P}} and \textit{\textbf{Q}}.
}
\label{fig:manifold-operation}
\end{figure*}

The Euclidean GD-based methods can not be used to solve the optimization problem of~\eqref{eqn:mani_optimization}, because they do not calculate the derivatives of the loss function with respect to the manifold geometry. Moreover, any translation between the current and next points on the gradient descent direction should also respect the geometry of Riemannian spaces.
To address these challenges, a commonly adopted solution involves the utilization of gradient-based Riemannian optimization techniques that benefit from the intrinsic geometry of Riemannian manifolds through the application of various Riemannian operations such as orthogonal projection, retraction, and parallel transport. The diagram presented in Fig. \ref{fig:manifold-operation} offers an elaborate visualization of these manifold operations, which are further elucidated in the subsequent discussions. The employment of these sophisticated mathematical tools enables facilitating the optimization process and enhancing the convergence toward optimal solutions.

\begin{definition}[Orthogonal projection]
For any point $\textit{\textbf{P}}\in \Manifold$, the orthogonal projection operator $\pi_\textit{\textbf{P}}:\mathbb{R}^n\rightarrow T_\textit{\textbf{P}}\Manifold$ transforms an Euclidean gradient vector to the Riemmanian counterpart (a vector located on the tangent space $T_\textit{\textbf{P}}\Manifold$).
\end{definition}
\begin{definition}[Exponential map and Retraction]
An exponential map, $\textrm{Exp}_{\textit{\textbf{P}}}(\cdot)$, maps a tangent vector $\textit{\textbf{v}}\in T_{\textit{\textbf{P}}} \Manifold$ to a manifold $\Manifold$. $\textrm{Exp}_{\textit{\textbf{P}}}(t \textit{\textbf{v}})$ represents the corresponding geodesic $\gamma(t):t\in [0,1]$, such that $\gamma(0) = \textit{\textbf{P}}$, $\dot{\gamma} (0) = \textit{\textbf{v}}$ on the manifold. However, evaluating the exponential map is computationally expensive. Retraction operation $R_{\textit{\textbf{P}}}:T_{\textit{\textbf{P}}} \Manifold \rightarrow \Manifold$ is usually used as a more efficient alternative that provides an estimation of exponential map.     
\end{definition}
\begin{definition}[Parallel transport]
The parallel transport operation $\Gamma_{\textit{\textbf{P}} \rightarrow \textit{\textbf{Q}}} : T_{\textit{\textbf{P}}} \Manifold \rightarrow T_{\textit{\textbf{Q}}} \Manifold$, takes a tangent vector $\textit{\textbf{v}} \in T_{\textit{\textbf{P}}} \Manifold$ of a point $\textit{\textbf{P}}$ and translates it to the tangent space of another point $\textit{\textbf{Q}}$ along a geodesic curve that connects $\textit{\textbf{P}}$ and $\textit{\textbf{Q}}$ on the manifold $\Manifold$ and outputs $\textit{\textbf{u}} \in T_{\textit{\textbf{Q}}} \Manifold$. Parallel transport preserves the norm of the vector during this process and ensures that the intermediate vectors are always lying on a tangent space. In practice,
since $\Gamma_{\textit{\textbf{P}}\rightarrow \textit{\textbf{Q}}}$ is unknown and incurs a high computational cost, one can use the orthogonal projection, as an approximation of mapping tangent vectors between $\textit{\textbf{P}}$ and $\textit{\textbf{Q}}$ on $\Manifold$ \cite{boothby1986introduction}. In this case, $\Gamma_{\textit{\textbf{P}}\rightarrow \textit{\textbf{Q}}}$ 
 for a given tangent vector $\textit{\textbf{v}} \in T_{\textit{\textbf{P}}} \Manifold$ is evaluated as $\pi_\textit{\textbf{Q}}(\textit{\textbf{v}})$ \cite{boothby1986introduction}.  
\end{definition}

Based on the above operations, $\textit{\textbf{P}}$ in Eqn.~\eqref{eqn:mani_optimization} is updated by:
\begin{equation}
\label{eqn:optim-update}
\begin{aligned}
\boldsymbol{P}^{(t+1)}&=R_{\boldsymbol{P}^{(t)}}\big(-\alpha \pi_{\boldsymbol{P}^{(t)}}(\nabla J)\big)
\end{aligned}
\end{equation}
where $\alpha$ is the step size and $-\alpha \pi_{\boldsymbol{P}^{(t)}}(\nabla J)$ is the search direction on the tangent space. Riemannian gradient descent algorithms compute search directions by transforming the Euclidean gradient $\nabla J$ to the Riemannian gradient using the orthogonal projection operation. After obtaining the Riemannian gradient on the tangent space, the retraction operation is applied to find the updated Riemannian parameter on manifolds. In other words, the retraction operation is analogous to traversing the manifold along the search direction.
In certain optimization approaches like Momentum Stochastic Gradient Descent (M-SGD) \citep{qian1999momentum}, accumulating gradient vectors throughout the optimization process is necessary. Nonetheless, the aggregation of gradients over time poses a challenge when dealing with a Riemannian manifold, unless the gradients are parallel transported to a shared tangent space.

\begin{table*}[]
\centering
\caption{The required Riemannian operators for Stiefel manifold, where $\text{Sym}(\textit{\textbf{P}})=\frac{1}{2}(\textit{\textbf{P}}+\textit{\textbf{P}}^\top)$ and $\text{qf}(\textit{\textbf{X}})$ is the $Q$ factor of the $QR$ decomposition of $X$.}
\label{tab:manifold operations}
\begin{tabular}{|c|c|c|}
\hline

Orthogonal projection $\pi_\textit{\textbf{P}}(\textit{\textbf{u}})$ & Retraction $R_{\textit{\textbf{P}}}(\textit{\textbf{v}})$ & Parallel transport  $\Gamma_{\textit{\textbf{P}} \rightarrow \textit{\textbf{y}}}(\textit{\textbf{w}})$ \\ 
\hline

$\textit{\textbf{u}}-\textit{\textbf{P}} \text{Sym}(\textit{\textbf{P}}^\top\textit{\textbf{u}})$ & $\text{qf}(\textit{\textbf{P}}+\textit{\textbf{v}})$ & $\pi_\textit{\textbf{y}}(\textit{\textbf{w}})$ \\
\hline
\end{tabular}
\end{table*}

\section{Proposed Method}
\label{Sec-method}
Generally, the optimization-based meta-training (learn to optimize) problem on Riemannian manifolds can be formulated as \cite{tabealhojeh2023rmaml}:
\begin{align}
     \label{Eq:general-Riemannian-meta-train}
\bm{\Theta}_{meta}^*= \underbrace{\min_{\bm{\Theta}\in\Manifold}{\frac{1}{N} \sum_{i=1}^{N}{\Loss_{i}\big(\bm{\Dataset}_i^{q},\overbrace{\bm{\Phi}_{i}^{*}}^{\text{inner-level}})}}}_{\text{outer-level}}\big) 
\end{align}
where $N$ is the number of training tasks, $\bm{\Theta}$ and $\bm{\Phi}_{i}$ respectively indicate the meta-parameters and parameters of the base-learners that are located on a Riemannian manifold $\Manifold$, ~\textit{i.e} $\bm{\Theta}, \bm{\Phi}_{i}\in\Manifold$. $\Loss_i$ is the loss function of the $i^{th}$ task, whereas $\Dataset_i=\{\bm{\Dataset}_i^{s},\bm{\Dataset}_i^{q}\}$ demonstrates the meta-train dataset of $i^{th}$ task. The support set $\bm{\Dataset}_i^{s}$ and the query set $\bm{\Dataset}_i^{q}$ are respectively analogous to training and validation datasets in classic supervised training. The main difference between~\eqref{Eq:general-Riemannian-meta-train} and the Euclidean meta-learning problem is that each parameter and meta-parameter of the Riemannian meta-learning is a matrix, which is a point on the Riemannian manifold, whereas in Euclidean meta-learning, each parameter or meta-parameter is a vector in Euclidean space. Equation~\eqref{Eq:general-Riemannian-meta-train} demonstrates two levels of optimization. The outer loop (or the meta-optimizer) learns the optimal meta-parameters, ~\textit{i.e} $\bm{\Theta}_{meta}^*$, using query datasets from a distribution of $N$ training tasks and the optimal task-specific parameters $\bm{\Phi}_{i}^*$. However, as the $\bm{\Phi}_{i}^*$ is unknown, it can be solved as another optimization problem (the inner-level optimization). Therefore,~\eqref{Eq:general-Riemannian-meta-train} can be considered as a bilevel optimization; a constrained optimization in which each constraint is itself an optimization problem:
\begin{align}
    \label{Eq:bi-level-Riemannian}
    \begin{split}
\bm{\Theta}_{meta}^*=\min_{\bm{\Theta}\in\Manifold}{\frac{1}{N} \sum_{i=1}^{N}{\Loss_{i}\big(\bm{\Dataset}_i^{q},\bm{\Phi}_i^*\big)}}
 \\ 
 \text{subject to}~~\bm{\Phi}_i^*=\gfun(\bm{\Dataset}_i^{s},\bm{\Theta}_{meta}^*)
\end{split}
\end{align}
where $\gfun(\bm{\Dataset}_i^{s},\bm{\Theta}_{meta}^*)$ is (usually) a gradient descent optimizer which is dependent on the meta-knowledge, $\bm{\Theta}_{meta}^*$. Assuming the meta-parameters as an initial point, the bilevel optimization problem of~\eqref{Eq:bi-level-Riemannian}, can be solved by performing two nested loops. At the inner loop, a limited number of gradient descent steps are performed to optimize each task-specific parameter $\bm{\Phi}_i^*$ using the meta-parameter as an initial point,~\textit{i.e} $\bm{\Phi}_i^{(0)}=\bm{\Theta}_{meta}$, and the outer loop performs a gradient descent update for the meta-parameter, $\bm{\Theta}_{meta}$.  Each Riemannian optimization step of the inner loop is as follows: 

\begin{equation}
\begin{aligned}
\label{eqn:task_specific_params_rmaml}
\bm{\Phi}_i^{(l+1)}=
R_{\bm{\Phi}_i^{(l)}}\left(-\alpha \pi_{\bm{\Phi}_i^{(l)}}\big(\nabla_{\bm{\Phi}_i^{(l)}} \Loss_i(\bm{\Dataset}_i^{s},\bm{\Phi}_i^{(l)})\big)\right), l\ge 1
\end{aligned}    
\end{equation}
Here, the gradient with respect to Euclidean space $\nabla_{\bm{\Phi}_i^{(l)}} \Loss_i(\bm{\Dataset}_i^{s},\bm{\Phi}_i^{(l)})$ is projected onto the tangent space of $\bm{\Phi}_i^{(l)}$, following which the parameters undergo an update using the \textit{retraction} operator {$R_{\bm{\Phi}_i^{(l)}}$}.

Moreover, a single step of gradient descent in Riemannian space for the outer loop is as follows:
\begin{align}
\label{eqn:rmaml_1}
\bm{\Theta}^{(t+1)}= R_{\bm{\Theta}^{(t)}}\left(-\beta \sum_{i=1}^{N}{\pi_{\bm{\Theta}^{(t)}}}\Big(\nabla_{\bm{\Theta}^{(t)}} \Loss_i\left(\bm{\Dataset}_i^{q},\bm{\Phi}_{i}^{*}\right)\Big)\right)     
\end{align}
 To determine $\bm{\Theta}^{(t+1)}$, the gradient vector in Euclidean space for each task, denoted as $\Big(\nabla_{\bm{\Theta}^{(t)}} \Loss_i\left(\bm{\Dataset}_i^{q},\bm{\Phi}_{i}^{*}\right)\Big)$, is projected orthogonally onto the tangent space of the current meta-parameter using the $\pi_{\bm{\Theta}^{(t)}}(\cdot)$ operation. Then, the summation over the Riemannian gradient of all the tasks is retracted to the manifold. To compute this gradient for each task, differentiating through the inner-level optimization path is needed. In other words, calculating $\pi_{\bm{\Theta}^{(t)}}\Big(\nabla_{\bm{\Theta}^{(t)}} \Loss_i\left(\bm{\Dataset}_i^{q},\bm{\Phi}_{i}^{*}\right)\Big)$ requires computing a gradient of the gradients or second order derivatives which has a high cost and includes backward computations through the Riemannian operators such as retraction and orthogonal projection. Assuming that $k$ Riemannian gradient descent steps are performed in the inner level optimization, then: 

\begin{align}
\label{eqn:nabla_derivatives}
\nabla_{\bm{\Theta}^{(t)}} \Loss_i\left(\bm{\Dataset}_i^{q},\bm{\Phi}_{i}^{*}\right)= \nabla_{\bm{\Theta}^{(t)}} \Loss_i\left(\bm{\Dataset}_i^{q},\bm{\Phi}_i^{(k)}\right)     
\end{align}

where $\bm{\Phi}_i^{(k)}$ is the result of inner-level optimization chain:
\begin{multline}
\label{eqn:inner_chain}
\bm{\Phi}_i^{(k)}=
R_{\bm{\Phi}_i^{(k-1)}}\left(-\alpha \pi_{\bm{\Phi}_i^{(k-1)}}\left(\nabla_{\bm{\Phi}_i^{(k-1)}} \Loss_i(\bm{\Dataset}_i^{s},\bm{\Phi}_i^{(k-1)})\right)\right)  \\  
\bm{\Phi}_i^{(k-1)}=
R_{\bm{\Phi}_i^{(k-2)}}\left(-\alpha \pi_{\bm{\Phi}_i^{(k-2)}}\left(\nabla_{\bm{\Phi}_i^{(k-2)}} \Loss_i(\bm{\Dataset}_i^{s},\bm{\Phi}_i^{(k-2)})\right)\right)    \\    
\vdots \\
\bm{\Phi}_i^{(0)}=\bm{\Theta}^{(t)}  
\end{multline}

Using the chain rule, the derivative in ~\eqref{eqn:nabla_derivatives} becomes as follows:

\begin{align}
\label{eqn:nabla_derivation_unfolded1}
\nabla_{\bm{\Theta}^{(t)}} \Loss_i\left(\bm{\Dataset}_i^{q},\bm{\Phi}_i^{(k)}\right)
= \frac{\partial \Loss_i}{\partial \bm{\Phi}_i^{(k)}}\frac{\partial \bm{\Phi}_i^{(k)}}{\partial \bm{\Theta}^{(t)}}
\end{align}

Considering the inner-level chain of the~\eqref{eqn:nabla_derivation_unfolded1}, the $\frac{\partial \bm{\Phi}_i^{(k)}}{\partial \bm{\Theta}^{(t)}}$ term can be further unfolded as follows:

\begin{multline}
\label{eqn:nabla_derivation_unfolded2}
\nabla_{\bm{\Theta}^{(t)}} \Loss_i\left(\bm{\Dataset}_i^{q},\bm{\Phi}_i^{(k)}\right)
= \frac{\partial \Loss_{i}\left(\bm{\Dataset}_i^{q},\bm{\Phi}_i^{(k)}\right)}{\partial \bm{\Phi}_i^{(k)}} \prod_{l=1}^{k}{  \frac{\partial \bm{\Phi}_i^{(l)}}{\partial \bm{\Phi}_{i}^{(l-1)}}}\\
= \frac{\partial \Loss_{i}\left(\bm{\Dataset}_i^{q},\bm{\Phi}_i^{(k)}\right)}{\partial \bm{\Phi}_i^{(k)}}\\
\prod_{l=1}^{k}{\frac{\partial \bigg( R_{\bm{\Phi}_i^{(l-1)}}\left(-\alpha \pi_{\bm{\Phi}_i^{(l-1)}}\left(\nabla_{\bm{\Phi}_i^{(l-1)}} \Loss_i(\bm{\Dataset}_i^{s},\bm{\Phi}_i^{(l-1)})\right)\right)\bigg)}{\partial \bm{\Phi}_{i}^{(l-1)}}}
\end{multline}

This equation contains the second-order differentiation terms that pass through the Riemannian operations of the inner-level optimization steps, which requires heavy calculation and computational resources. 

In the next section, we formulate a first-order approximation of ~\eqref{eqn:nabla_derivation_unfolded2}, that enables us to perform a Hessian-free bi-level optimization for this Riemannian meta-learning problem. 

\subsection{First order approximation: from Euclidean space to Stiefel manifold}
Without loss of generality, let us consider exactly one update step at the inner-level chain of~\eqref{eqn:nabla_derivation_unfolded2}. Thus, this step of inner-level optimization will be as follows:

\begin{align}
\label{eqn:inner_single_step}
    \bm{\Phi}_i^{(1)}=
R_{\bm{\Theta}^{(t)}}\left(-\alpha \pi_{\bm{\Theta}^{(t)}}\left(\nabla_{\bm{\Theta}^{(t)}} \Loss_i(\bm{\Dataset}_i^{s},\bm{\Theta}_i^{(t)})\right)\right)
\end{align}

Therefore, equation \eqref{eqn:nabla_derivation_unfolded2} will be reduced to:
\begin{multline}
    \label{eqn:nabla_derivatives_single}
    \nabla_{\bm{\Theta}^{(t)}} \Loss_i\left(\bm{\Dataset}_i^{q},\bm{\Phi}_i^{(k)}\right)=\\
    \frac{\partial \Loss_{i}\left(\bm{\Dataset}_i^{q},\bm{\Phi}_i^{(1)}\right)}{\partial \bm{\Phi}_i^{(1)}} \frac{\partial \bigg( R_{\bm{\Theta}^{(t)}}\left(-\alpha \pi_{\bm{\Theta}^{(t)}}\left(\nabla_{\bm{\Theta}^{(t)}} \Loss_i(\bm{\Dataset}_i^{s},\bm{\Theta}^{(t)})\right)\right)\bigg)}{\partial \bm{\Theta}^{(t)}}
\end{multline}

The above expression applies for all Riemannian manifolds. For instance, using the Euclidian operations on the Euclidean manifold, one can write~\eqref{eqn:nabla_derivatives_single} as:
\begin{multline}
    \label{eqn:nabla_derivatives_single_Euc}
    \nabla_{\bm{\theta}^{(t)}} \Loss_i\left(\bm{\Dataset}_i^{q},\bm{\phi}_i^{(k)}\right)=\\
    \frac{\partial \Loss_{i}\left(\bm{\Dataset}_i^{q},\bm{\phi}_i^{(1)}\right)}{\partial \bm{\phi}_i^{(1)}} 
    \frac{\partial \bigg( \bm{\theta}^{(t)}-\alpha (\nabla_{\bm{\theta}^{(t)}} \Loss_i(\bm{\Dataset}_i^{s},\bm{\theta}^{(t)})\bigg)}{\partial \bm{\theta}^{(t)}}
\end{multline}

which can be further simplified to:

\begin{multline}
\label{eqn:nabla_derivatives_single_Euc_sim}
    \nabla_{\bm{\theta}^{(t)}} \Loss_i\left(\bm{\Dataset}_i^{q},\bm{\phi}_i^{(k)}\right)=\\
    \frac{\partial \Loss_{i}\left(\bm{\Dataset}_i^{q},\bm{\phi}_i^{(1)}\right)}{\partial \bm{\phi}_i^{(1)}}  \bigg( \bm{I}-\alpha (\nabla^{2}_{\bm{\theta}^{(t)}} \Loss_i(\bm{\Dataset}_i^{s},\bm{\theta}^{(t)})\bigg)
\end{multline}

For the Euclidean spaces, the first-order approximation of~\eqref{eqn:nabla_derivatives_single_Euc_sim}, FOMAML \cite{finn2017model} can be calculated by ignoring its second-order derivatives:
\begin{align}
    \label{eqn:derivatives9}
    \nabla_{\bm{\theta}^{(t)}} \Loss_i\left(\bm{\Dataset}_i^{q},\bm{\phi}_i^{(k)}\right)
    =\frac{\partial \Loss_{i}\left(\bm{\Dataset}_i^{q},\bm{\phi}_i^{(1)}\right)}{\partial \bm{\phi}_i^{(1)}}
\end{align}

  On the other hand, calculating such an approximation for the Riemannian spaces is not straightforward. In the rest of this section, we derive the formulation of first-order approximation on Stiefel manifolds. For simplification, we indicate $\nabla_{\bm{\Theta}^{(t)}} \Loss_i(\bm{\Dataset}_i^{s},\bm{\Theta}^{(t)})$ as $\nabla_{\bm{\Theta}^{(t)}}\Loss_i$, and $d$ indicates the differential operator. Let us assume that $H^{'}=\frac{d \bigg( R_{\bm{\Theta}^{(t)}}\big(-\alpha \pi_{\bm{\Theta}^{(t)}}\left(\nabla_{\bm{\Theta}^{(t)}} \Loss_i(\bm{\Dataset}_i^{s},\bm{\Theta}^{(t)})\right)\big)\bigg)}{d (\bm{\Theta}^{(t)})}$, we aim to compute $H^{'}$ such that:
\begin{align}\label{eqn:differential}
   d \bigg( R_{\bm{\Theta}^{(t)}}\big(-\alpha \pi_{\bm{\Theta}^{(t)}}\left(\nabla_{\bm{\Theta}^{(t)}} \Loss_i\right)\big)\bigg)=H^{'}d (\bm{\Theta}^{(t)})
\end{align}
applying the differential on the left side of~\eqref{eqn:differential} and substituting $R_{\bm{P}}(\bm{v})=P+v$ (an approximated version of Stiefel retraction) in that equation results in:
\begin{multline}\label{eqn:esbat1}
d\bigg(R_{\bm{\Theta}^{(t)}}\left(-\alpha \pi_{\bm{\Theta}^{(t)}}\left(\nabla_{\bm{\Theta}^{(t)}} \Loss_i\right)\right)\bigg)=\\
   d\bigg( \bm{\Theta}^{(t)}-\alpha \pi_{\bm{\Theta}^{(t)}}\left(\nabla_{\bm{\Theta}^{(t)}} \Loss_i\right) \bigg)=\\
   d(\bm{\Theta}^{(t)})-d\bigg(\alpha \pi_{\bm{\Theta}^{(t)}}\left(\nabla_{\bm{\Theta}^{(t)}} \Loss_i\right) \bigg)
\end{multline}
 considering the Stiefel orthogonal projection $\pi_{\bm{P}}\left(\bm{v}\right)$ from Table~\ref{tab:manifold operations}, equation~\eqref{eqn:esbat1} becomes as follows:    
\begin{multline}\label{eqn:esbat2}
        d(\bm{\Theta}^{(t)})-\alpha d\bigg( \nabla_{\bm{\Theta}^{(t)}}\Loss_i-\bm{\Theta}^{(t)}\frac{(\bm{\Theta}^{(t)})^{T} \nabla_{\bm{\Theta}^{(t)}}\Loss_i-(\nabla_{\bm{\Theta}^{(t)}}\Loss_i)^{T}\bm{\Theta}^{(t)}}{2} \bigg)\\
    =d(\bm{\Theta}^{(t)})-\alpha d(\nabla_{\bm{\Theta}^{(t)}}\Loss_i)+0.5\alpha d\bigg( \bm{\Theta}^{(t)} (\bm{\Theta}^{(t)})^{T} \nabla_{\bm{\Theta}^{(t)}}\Loss_i\bigg)\\-0.5\alpha d\bigg(\bm{\Theta}^{(t)}(\nabla_{\bm{\Theta}^{(t)}}\Loss_i)^{T}\bm{\Theta}^{(t)} \bigg)
\end{multline}
considering $\bm{\Theta}^{(t)} (\bm{\Theta}^{(t)})^{T}=\bm{I}$ for orthogonal matrices, the third part of equation~\eqref{eqn:esbat2} can be further simplified, which results in:  
\begin{multline}\label{eqn:esbat3}
 d(\bm{\Theta}^{(t)})-\alpha d(\nabla_{\bm{\Theta}^{(t)}}\Loss_i)+0.5\alpha d(\nabla_{\bm{\Theta}^{(t)}}\Loss_i)\\-0.5\alpha d\bigg(\bm{\Theta}^{(t)}(\nabla_{\bm{\Theta}^{(t)}}\Loss_i)^{T}\bm{\Theta}^{(t)} \bigg)\\=
    d(\bm{\Theta}^{(t)})-0.5\alpha d(\nabla_{\bm{\Theta}^{(t)}}\Loss_i)-0.5\alpha d\bigg(\bm{\Theta}^{(t)}(\nabla_{\bm{\Theta}^{(t)}}\Loss_i)^{T}\bm{\Theta}^{(t)} \bigg)
\end{multline}
using the relation $d(XYZ)=d(X)YZ+Xd(Y)Z+XYd(Z)$ for the third part of~\eqref{eqn:esbat3}, it becomes as follows:  
\begin{multline}
    d(\bm{\Theta}^{(t)})-0.5\alpha d(\nabla_{\bm{\Theta}^{(t)}}\Loss_i)
    -0.5\alpha d(\bm{\Theta}^{(t)})(\nabla_{\bm{\Theta}^{(t)}}\Loss_i)^{T}\bm{\Theta}^{(t)}\\-0.5\alpha \bm{\Theta}^{(t)}d(\nabla_{\bm{\Theta}^{(t)}}\Loss_i)^{T}\bm{\Theta}^{(t)}-0.5\alpha \bm{\Theta}^{(t)}(\nabla_{\bm{\Theta}^{(t)}}\Loss_i)^{T}d(\bm{\Theta}^{(t)})
\end{multline}
The second and fourth terms of the above formula contain second-order derivatives. By ignoring those terms and using $\vect(\cdot)$ operator, the derivatives with respect to $\bm{\Theta}$ which is an orthogonal matrix (a point on Stiefel manifold) can be written as:
\begin{multline}
 \label{eqn:derivatives11}
  \vect\big( d(\bm{\Theta}^{(t)})\big)-0.5\alpha \vect\big( d(\bm{\Theta}^{(t)})(\nabla_{\bm{\Theta}^{(t)}}\Loss_i)^{T}\bm{\Theta}^{(t)}\big)\\-0.5\alpha \vect\big(\bm{\Theta}^{(t)}(\nabla_{\bm{\Theta}^{(t)}}\Loss_i)^{T}d(\bm{\Theta}^{(t)})\big)   
\end{multline}
knowing that $vec(AXB)=(B^{T}\otimes A)vec(X)$, where $\otimes$ indicates the Kronecker product, equation~\eqref{eqn:derivatives11} becomes:
 \begin{multline}\label{eqn:derivatives11-a}
  \bm{I}\vect\big( d(\bm{\Theta}^{(t)})\big)-0.5\alpha\bigg((\bm{\Theta}^{(t)})^{T}\nabla_{\bm{\Theta}^{(t)}}\Loss_i\otimes \bm{I}\bigg) \vect( d(\bm{\Theta}^{(t)}))\\
  -0.5\alpha\bigg(\bm{I}\otimes \bm{\Theta}^{(t)}(\nabla_{\bm{\Theta}^{(t)}}\Loss_i)^{T} \bigg) \vect( d(\bm{\Theta}^{(t)}))
\end{multline}
using $(A\otimes I)+(I\otimes B)=A\oplus B$, equation~\eqref{eqn:derivatives11-a} can be rewritten as follows:
\begin{multline}
    \bigg(\bm{I}-0.5\alpha\big((\bm{\Theta}^{(t)})^{T}\nabla_{\bm{\Theta}^{(t)}}\Loss_i\oplus \bm{\Theta}^{(t)}(\nabla_{\bm{\Theta}^{(t)}}\Loss_i)^{T} \big)\bigg)\vect( d(\bm{\Theta}^{(t)}))  
\end{multline}
in which the symbol $\oplus$ indicates the Kronecker sum. Thus, we will have:
\begin{multline}
    \label{eqn:derivatives12}
    \vect\bigg(d\bigg(R_{\bm{\Theta}^{(t)}}\left(-\alpha \pi_{\bm{\Theta}^{(t)}}\left(\nabla_{\bm{\Theta}^{(t)}} \Loss_i(\bm{\Dataset}_i^{s},\bm{\Theta}^{(t)})\right)\right)\bigg)\bigg)\approx \\
    \bigg(\bm{I}-0.5\alpha\big((\bm{\Theta}^{(t)})^{T}\nabla_{\bm{\Theta}^{(t)}}\Loss_i\oplus \bm{\Theta}^{(t)}(\nabla_{\bm{\Theta}^{(t)}}\Loss_i)^{T} \big)\bigg)\big(\vect( d(\bm{\Theta}^{(t)})\big)
\end{multline}
As a result:
\begin{multline}
 H^{'}=\bm{I}-0.5\alpha\big((\bm{\Theta}^{(t)})^{T}\nabla_{\bm{\Theta}^{(t)}}\Loss_i\oplus \bm{\Theta}^{(t)}(\nabla_{\bm{\Theta}^{(t)}}\Loss_i)^{T} \big)   
\end{multline}
and the first-order approximation of~\eqref{eqn:nabla_derivatives_single} is as follows:
\begin{multline}   
    \label{eqn:final_FORML}
    \nabla_{\bm{\Theta}^{(t)}} \Loss_i\left(\bm{\Dataset}_i^{q},\bm{\Phi}_i^{(k)}\right)
    \approx\frac{\partial \Loss_{i}\left(\bm{\Dataset}_i^{q},\bm{\Phi}_i^{(1)}\right)}{\partial \bm{\Phi}_i^{(1)}}\\
    \bigg(\bm{I}-0.5\alpha\big((\bm{\Theta}^{(t)})^{T}\nabla_{\bm{\Theta}^{(t)}}\Loss_{i}\oplus \bm{\Theta}^{(t)}(\nabla_{\bm{\Theta}^{(t)}}\Loss_{i})^{T} \big) \bigg)
\end{multline}

Equation~\eqref{eqn:final_FORML} computes the derivatives of the outer loop without the need for calculating the Hessian matrices. Moreover, the approximation consists of the Kronecker sum of a matrix with its transpose, which further saves the calculations' complexity cost.    

\subsection{Generalization to $k$ inner-level optimization steps}
In the previous section, we formulated a first-order approximation for the bi-level meta-learning problem, assuming just one inner-level optimization step. Using~\eqref{eqn:final_FORML}, it is straightforward to formulate the first-order approximation of~\eqref{eqn:nabla_derivation_unfolded2} as follows:
\begin{multline}
\label{eqn:final_FORML-k}
\nabla_{\bm{\Theta}^{(t)}} \Loss_i\left(\bm{\Dataset}_i^{q},\bm{\Phi}_i^{(k)}\right)
\approx \frac{\partial \Loss_{i}\left(\bm{\Dataset}_i^{q},\bm{\Phi}_i^{(k)}\right)}{\partial \bm{\Phi}_i^{(k)}}
\\\prod_{l=1}^{k}{\bigg(\bm{I}-0.5\alpha\big((\bm{\Phi}_{i}^{(l-1)})^{T}\nabla_{\bm{\Phi}_{i}^{(l-1)}}\Loss_{i}\oplus \bm{\Phi}_{i}^{(l-1)}(\nabla_{\bm{\Phi}_{i}^{(l-1)}}\Loss_{i})^{T} \big) \bigg)}
\end{multline}

\begin{figure}[t]
\centering
\includegraphics[scale=0.25, keepaspectratio]{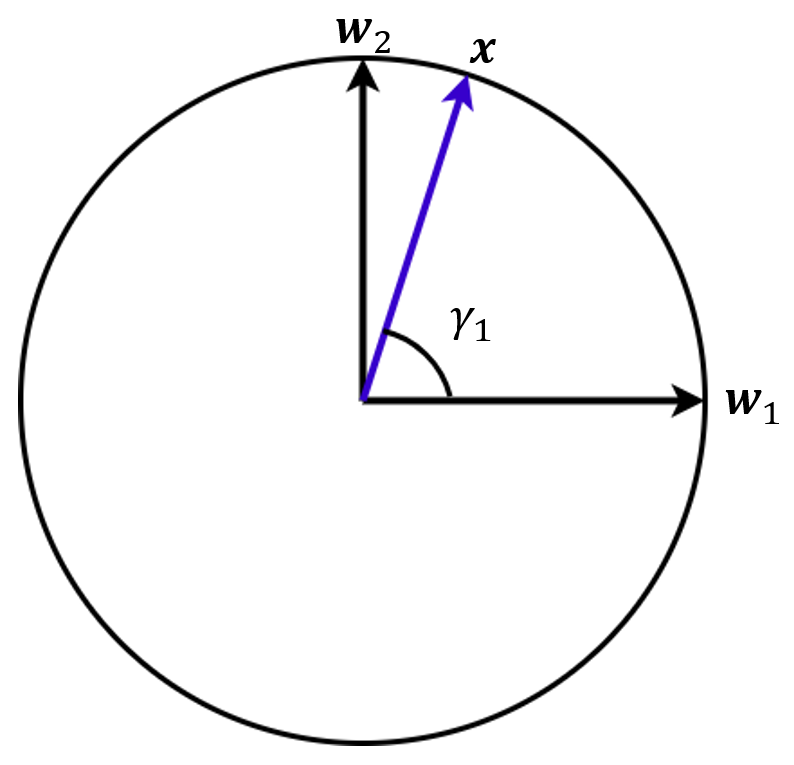}
\caption{A sample representation of the Stiefel fully connected layer for 2D output space, where $\bm{W}=[\bm{w}_1,\bm{w}_2]$ represents the orthogonal weight matrix (lies on Stiefel manifold) and $\bm{x}$ is the input vector of the Stiefel fully connected layer. For this example, equation~(\ref{eqn:Stiefel-layer}) becomes: $\bm{W}^{T}\bm{x}=\bm{\gamma}=[\gamma_1,\gamma_2]$.}
\label{fig:Stiefellayer}
\end{figure} 
\subsection{The proposed bi-level algorithm}
The full proposed method, called FORML, is outlined in Algorithm~\ref{alg-RMAML}. Practically we meta-optimize the last fully-connected layer of the backbone network using FORML on Stiefel manifold, while the other parameters are optimized on Euclidean space. Moreover, we normalize the input of the Stiefel layer. As a result, the forward computation of the fully-connected layer acts as a cosine similarity between the input vector $\textbf{x}$ and the weight vectors of the classes that are columns of the weight matrix $\textbf{W}$ (as shown in Fig.~\ref{fig:Stiefellayer}).  
\begin{align}
    \label{eqn:Stiefel-layer}
    \textbf{W}^{T}\textbf{x}=\frac{\textbf{W}^{T}\textbf{x}}{\| \textbf{W} \| \| \textbf{x} \|}=\bm{\gamma}
\end{align}

\begin{algorithm*}
\centering
\caption{FORML}\label{alg-RMAML}
\begin{algorithmic}[1]
\Require The distribution of meta-training tasks $p(\Tau)$.
\vspace{0.1cm}
\Require Learning rate of the inner-level loop $\alpha$, Learning rate of the outer-level loop $\beta$.
\vspace{0.1cm}
\State Initialize the meta-parameters $\bm{\Theta}^{(0)}$ randomly. 
\vspace{0.1cm}
\While{not done}
    \vspace{0.1cm}
    \State Sample batch $\{\Tau_i\}_{i=1}^B$ 
    \vspace{0.1cm}
    \For {all $\Tau_i$}
        \vspace{0.1cm}
        \State Sample $k$ data points $\bm{\Dataset}_i^{s}=\{(\textit{\textbf{x}}_j,\textit{\textbf{y}}_j)\}_{j=1}^k$ from $\Tau_i$ for the inner-level updates
        \vspace{0.1cm}
        
        \State Calculate $\nabla_{\bm{\Phi}_i^{(l)}} \Loss_i(\bm{\Dataset}_i^{s},\bm{\Phi}_i^{(l)})$ with respect to $k$ samples.
        \vspace{0.1cm}
        \State Compute the task-specific parameters with gradient descent using~\eqref{eqn:task_specific_params_rmaml}.
        \vspace{0.1cm}
    \EndFor
    \vspace{0.1cm}
    \State Sample $m$ data points $\bm{\Dataset}_i^{q}=\{(\textit{\textbf{x}}_j,\textit{\textbf{y}}_j)\}_{j=1}^m$ from $\Tau_i$ for the meta-level (outer-level) update
        \vspace{0.1cm}
    \State  Compute $\nabla_{\bm{\Theta}^{(t)}} \Loss_i\left(\bm{\Dataset}_i^{q},\bm{\Phi}_i^{(k)}\right)$ using first-order approximation method of~\eqref{eqn:final_FORML-k}.
    \vspace{0.1cm}
    \State Compute $\bm{\Theta}^{(t+1)}$ using~\eqref{eqn:rmaml_1}.
    \vspace{0.1cm}
\EndWhile
\end{algorithmic}
\end{algorithm*}

\section{Experimental Results}
\label{Sec-Results}

\subsection{Evaluation Scenarios and Datasets}
Following the experimental setup outlined by \cite{tabealhojeh2023rmaml,chen2019closer}, we assess our proposed approach across various few-shot classification tasks, including single-domain and cross-domain scenarios. Within the single-domain experiments, the meta-train, meta-validation, and meta-test tasks are drawn from a single dataset. In contrast, in the cross-domain scenarios, the meta-test tasks originate from a distinct dataset. Consequently, the impact of domain shift on our proposed approach is investigated within the context of cross-domain experiments.

Two experiments are carried out to tackle the single-domain scenario: object recognition and fine-grained image classification. Object recognition is performed using the Mini-ImageNet dataset \citep{ravi2016optimization}, a subset of 100 classes from the ImageNet dataset, with 600 images sampled from each class. As per the setting provided by Ravi and Larochelle \citep{ravi2016optimization}, the dataset is randomly partitioned into 64, 16, and 20 classes for meta-train, meta-validation, and meta-test sets respectively.

The CUB-200-2011 (CUB) and FC100 datasets are utilized for fine-grained image classification. The CUB dataset comprises 11,788 images of 200 bird classes, segmented into train, validation, and test meta-sets with 100, 50, and 50 classes respectively \cite{krizhevsky2009learning}. The FC100 dataset, derived from CIFAR100, presents a more demanding few-shot scenario due to lower image resolution and intricate meta-training/test splits based on object superclasses. With 100 classes and 600 samples per class of 32×32 color images, the dataset is divided into 20 superclasses. Krizhevsky et al.'s proposed splits \cite{krizhevsky2009learning}, allocate 60 classes from 12 super-classes for meta-training, while meta-validation and meta-test sets consist of 20 classes from 4 super-classes, respectively. This separation by super-classes serves to reduce information overlap between training and validation/test tasks.

One crucial element of meta-learning involves analyzing the impact of domain shift between the meta-training and meta-testing datasets on the meta-learning algorithms. An example would be the noticeable difference in domain gap between the meta-training and meta-testing categories of miniImageNet compared to CUB within the word-net hierarchy \citep{miller1995wordnet}. Consequently, the task of meta-learning on miniImageNet poses greater challenges compared to the CUB dataset. Thus, our experimental setup also delves into cross-domain scenarios. Through the cross-domain experiment, we aim to explore how domain shift influences our proposed approach, despite its original design for single-domain classification.
Similar to \citep{chen2019closer}, we carry out cross-domain experiments such as \emph{mini-ImageNet$\rightarrow$CUB} and \emph{Omniglot$\rightarrow$EMNIST}. In the \emph{mini-ImageNet$\rightarrow$CUB} experiment, all 100 classes from mini-ImageNet are utilized as the meta-training set, with 50 meta-validation and 50 meta-testing classes randomly selected from CUB. As for the \emph{Omniglot$\rightarrow$EMNIST} experiment, we adopt the framework proposed by Dong and Xing \citep{dong2018domain}, where the non-Latin characters from the Omniglot dataset serve as the meta-training classes, and the EMNIST dataset \citep{cohen2017emnist}, containing 10 digits and upper and lower case English alphabets, is allocated for meta-validation and meta-testing classes. The meta-training set comprises a total of 1597 classes from the Omniglot dataset, while the 62 EMNIST classes are divided into 31 for meta-validation and 31 for meta-testing sets.

\subsection{Experimental Details}
We employ the standardized $N\text{-way}$ $k\text{-shot}$ classification protocol, where each task instance comprises randomly chosen samples from $N$ distinct classes of the meta-training set. For every class, $k$ and $q$ instances are sampled for the support and query sets, respectively. During the meta-testing phase, we conduct 600 trials and present the mean classification accuracy and the 95\% confidence interval of these 600 experiments as the final accuracy. $N$, $k$, and $q$ are chosen similarly to the meta-training stage and $N$ remains constant at 5 for all experiments across various datasets.

In the case of mini-ImageNet and CUB datasets, we utilize convolutional backbones with four and six layers (Conv-4 and Conv-6) having an input dimension of $84\times 84$. Additionally, we incorporate ResNet-10 and ResNet-18 backbone networks with an input size of $224\times 224$. For the FC100 experiment, Conv-4 and ResNet-12 backbone networks are employed. Furthermore, Conv-4 is utilized for the cross-domain character recognition experiment (\emph{Omniglot$\rightarrow$EMNIST}), while the ResNet-18 network is used for the \emph{mini-ImageNet$\rightarrow$CUB} experiment.

In all experiments across diverse network backbones, it is assumed that the parameters of the final fully connected layer of the network are orthonormal and located on a Stiefel manifold. FORML learns this matrix utilizing Riemannian operators, as detailed in Algorithm~\ref{alg-RMAML}. FORML optimizes the parameters of the remaining layers in Euclidean space. All experiments were conducted on a single NVIDIA V100 GPU with 32GB of video memory.

\subsubsection{Hyper-parameter setups}
We indicate the learning rate of inner and outer loops for Stiefel layers as $\alpha$ and $\beta$ respectively, while for the Euclidean layers, the learning rate of the outer loop is shown by $\widehat\beta$. For \textbf{single-domain} scenarios : 
\begin{itemize} 
\item The parameter $\alpha$ is configured as $0.1$ for experiments involving Conv-4 and Conv-6 backbone architectures, while it is adjusted to $0.01$ for other cases. 
\item $\widehat\beta$ is established at $10^{-3}$ for all trials with Conv-4 and Conv-6 backbone designs. \item The value of $\widehat\beta$ is fixed at $10^{-4}$ for experiments employing ResNet backbone structures. 
\end{itemize}

Likewise, in the \textbf{cross-domain} setup: \begin{itemize} \item The parameter $\alpha$ is defined as $0.01$ for the mini-ImageNet$\rightarrow$CUB scenario and is adjusted to $0.05$ and $0.1$ for Omniglot$\rightarrow$EMNIST 1-shot and 5-shot experiments, respectively. 
\item For mini-ImageNet$\rightarrow$CUB, $\widehat\beta$ is set at $5\times10^{-4}$, while for Omniglot$\rightarrow$EMNIST 1-shot and 5-shot setups, the values are $10^{-3}$ and $10^{-4}$, respectively. 
\end{itemize}
The learning rate $\widetilde\beta$ in all experiments is initialized at $10^{-3}$. To have a fair comparison, the hyper-parameters are determined through trial and error, and consistently across all experiments, the batch size and the number of inner-level updates during training are set at 4 and 5, respectively.

 \begin{table*}
 \centering
 \caption{A comparison of few-shot classification accuracies between first-order approximation methods in Euclidean space (MAML) and Stiefel manifold (FORML) for mini-ImageNet and CUB datasets against different backbone networks, using 5-way classification setting.}\label{tab1: all experiments}
 \begin{tabular}{lllll}
 \hline
 \multicolumn{1}{c}{\textbf{Dataset}}  & \textbf{Method} & \textbf{Conv-4} & \textbf{Conv-6} & \textbf{ResNet-10} 
 \\ \hline
 \multirow{2}{*}{\begin{tabular}[l]{@{}l@{}}CUB 1-shot\end{tabular}} &  MAML (First-Order) \cite{chen2019closer}  & $54.73\pm0.97$ &  $66.26\pm1.05$& $70.32\pm0.99$ 
 \\
 & FORML (ours) & $\bm{58.86\pm0.99}$ & $\bm{67.29\pm0.99}$ & $\bm{70.96\pm0.97}$  
 \\
  \hline
  \multirow{2}{*}{\begin{tabular}[l]{@{}l@{}}CUB 5-shot\end{tabular}} & MAML (First-Order) \cite{chen2019closer} & $75.75\pm0.76$& $78.82\pm0.70$& $80.93\pm0.71$  
  \\
  & FORML (ours) & $\bm{77.93\pm0.95}$& $\bm{79.76\pm0.98}$7 & \bm{$82.45 \pm 0.69$}   
  \\
  \hline
  \multirow{2}{*}{\begin{tabular}[l]{@{}l@{}}mini-ImageNet \\ 1-shot\end{tabular}} 
  &MAML (First-Order)\cite{chen2019closer} &$46.47\pm0.82$ &$50.96\pm0.92$& $54.69\pm0.89$ 
  \\
 & FORML (ours) &\bm{$49.94\pm0.84$} & \bm{$51.68\pm0.55$}& \bm{$55.13 \pm 0.88$}  
 \\
  \hline
  \multirow{2}{*}{\begin{tabular}[l]{@{}l@{}}mini-ImageNet \\ 5-shot\end{tabular}} 
 & MAML (First-Order) \cite{chen2019closer} &$62.71\pm0.71$& $66.09\pm0.71$ &$66.62\pm0.83$ 
 \\
 & FORML (ours) & \bm{$65.46\pm0.68$} & \bm{$68.12\pm0.88$}& \bm{$68.82\pm0.85$} 
 \\
  \bottomrule
 \end{tabular}
 \end{table*}

\begin{table}
\centering
\caption{Few-shot classification accuracies for mini-ImageNet datasets against Conv-4 backbone network. 5-way classification setting is used in all experiments. The best results are shown in \textbf{bold}.}\label{tab1: Mini_Conv4_experiments}
\begin{tabular}{lll}
\hline
\multicolumn{1}{c}{\textbf{Method}}  &  \textbf{1-shot} & \textbf{5-shot} \\ \hline
MAML (first-0order) \cite{chen2019closer} &  $46.47\pm0.82$ & $62.71\pm0.71$\\
Reptile \cite{nichol2018first} &$47.07\pm0.26$ &$62.74\pm0.38$\\
ANIL \cite{raghu2019rapid}  & $47.20\pm0.27$  &$62.59\pm0.39$ \\
BOIL \cite{oh2021boil}  &$47.76\pm0.31$ &$64.40\pm0.28$ \\
RMAML \cite{tabealhojeh2023rmaml} & \bm{$50.03 \pm  0.84$} &  $65.12 \pm 0.76$\\
\hline
FORML (ours)  &\bm{$49.94\pm0.84$} &\bm{$65.46\pm0.68$}\\
 \bottomrule
\end{tabular}
\end{table}

\begin{table}[t]
\centering
\caption{Results of Few-shot classification results on CUB dataset using Conv-4 backbone.}\label{tab:further-exp-CUB}
\begin{tabular}{ccc}
\hline

\textbf{Method} & \textbf{1-shot} & \textbf{5-shot} \\ \hline
ANIL \cite{raghu2019rapid}  & $47.20\pm0.27$  &$62.59\pm0.39$ \\
BOIL \cite{oh2021boil}  &$47.76\pm0.31$ &$64.40\pm0.28$ \\
RMAML\cite{tabealhojeh2023rmaml} & \bm{$75.79 \pm  0.92$} &  \bm{$86.77 \pm 0.57$} \\
\hline
FORML (ours) &$\bm{74.86\pm0.92}$  &$86.35\pm0.78$ \\
\hline
\end{tabular}
\end{table}

\begin{table*}
\centering
\caption{Few-shot classification accuracies for FC100 dataset. 5-way classification setting is used in all experiments.}\label{tab:FC100-exp}
\begin{tabular}{lllll}
\toprule
\textbf{Backbone Networks} & \multicolumn{2}{c}{\textbf{Conv-4}} & \multicolumn{2}{c}{\textbf{ResNet-12}} \\ 
\cmidrule(r){1-5}
\textbf{Method} & \textbf{1-shot} & \textbf{5-shot} & \textbf{1-shot} & \textbf{5-shot} \\
\cmidrule(r){1-1}
\cmidrule(r){2-3}
\cmidrule(r){4-5}
Baseline & $32.48\pm 0.58$ & $47.15 \pm 0.67$ & $38.24 \pm 0.63$ & $56.87 \pm 0.71$ \\
MAML & $36.70\pm 0.83$ & $47.69\pm0.79$ & $40.15\pm0.78$ & $54.13\pm0.77$ \\
ANIL \cite{raghu2019rapid} &$36.37\pm0.33$  &$45.65\pm0.44$   & - &- \\
BOIL \cite{oh2021boil}    &$38.93\pm0.45$ &$51.66\pm0.32$&-&- \\
RMAML \cite{tabealhojeh2023rmaml} & \bm{$39.92 \pm 0.64$} &  \bm{$51.34 \pm 0.72$} & \bm{$42.58\pm 0.70$} & \bm{$59.68\pm 0.78$}  \\
\hline
FORML (ours) & \bm{$38.82\pm0.72$} & \bm{$49.05\pm0.88$} & \bm{$41.30\pm0.68$} & \bm{$57.35\pm0.84$}  \\ 
\bottomrule

\end{tabular}
\end{table*}

\begin{table}
\centering
\caption{Few-shot classification accuracies for mini-ImageNet datasets against Conv-4 backbone network. 5-way classification setting is used in all experiments.}\label{tab1: Mini_Conv4_experiments_sup}
\begin{tabular}{lll}
\hline
\multicolumn{1}{c}{\textbf{Method}}  &  \textbf{1-shot} & \textbf{5-shot} \\ \hline
ML-LSTM~\citep{Sachin2017} & $43.44\pm0.77$ & $60.60\pm0.71$ \\
iMAML-HF~\citep{rajeswaran2019meta} & $49.30\pm1.88$ & -  \\
Meta-Mixture~\citep{jerfel2019reconciling} &  $49.60\pm1.50$ & $64.60\pm0.92$ \\
Amortized VI~\citep{gordon2019meta} & $44.13\pm1.78$ & $55.68\pm0.91$ \\
DKT+BNCosSim~\citep{patacchiola2020bayesian} & $49.73\pm 0.07$ & $64.00\pm0.09$   \\
\hline
FORML (ours)  &\bm{$49.94\pm0.84$} &\bm{$65.46\pm0.68$}\\
 \bottomrule
\end{tabular}
\end{table}

 \begin{table}[t]
\centering
\caption{Comparison with other state-of-the-art methods for Few-shot classification on CUB dataset using ResNet-18 backbone. }\label{tab:further-exp-CUB2}
\begin{tabular}{ccc}
\hline

\textbf{Method} & \textbf{1-shot} & \textbf{5-shot} \\ \hline
Robust-20~\citep{dvornik2019diversity} & $58.67 \pm 0.70$& $75.62 \pm 0.50$ \\ 
GNN-LFT~\citep{tseng2020cross} & $51.51 \pm 0.80$& $73.11 \pm 0.68$ \\ 
Baseline++~\citep{chen2019closer} & $67.02\pm0.90$& $83.58\pm0.50$ \\
Arcmax~\citep{afrasiyabi2020associative} & $71.37\pm0.90$ & $85.74\pm0.50$ \\
MixtFSL~\citep{afrasiyabi2021mixture} & $73.94\pm1.10$ &$86.01\pm0.50$ \\
RMAML\cite{tabealhojeh2023rmaml} & \bm{$75.79 \pm  0.92$} &  \bm{$86.77 \pm 0.57$} \\
\hline
FORML (ours) &$\bm{74.86\pm0.92}$  &$86.35\pm0.78$ \\
\hline
\end{tabular}
\end{table}

\begin{table*}
\centering
\caption{Supplementary comparison for few-shot classification results on FC100 dataset.}\label{tab:further-exp-FC100}
\begin{tabular}{cccc}
\hline
\textbf{Method} & \textbf{Backbone} & \textbf{1-shot} & \textbf{5-shot} \\ \hline
MetaOptNet~\citep{lee2019meta} & ResNet-12 & $41.10\pm0.60$& $55.50\pm0.60$ \\
Arcmax~\citep{afrasiyabi2020associative} & ResNet-18 & $40.84\pm0.71$ & $57.02\pm0.63$ \\
MixtFSL~\citep{afrasiyabi2021mixture} & ResNet-18 & \bm{$41.50\pm0.67$} & \bm{$58.39\pm0.62$} \\
FORML (ours) & ResNet-12 & $41.30\pm0.68$ & $57.35\pm0.84$\\ \hline

\end{tabular}
\end{table*}
\subsection{Results for Single-Domain Experiments}

We examined our proposed method against single-domain few-shot classification benchmarks including mini-ImageNet, CUB, and FC100 benchmarks. All experiments are conducted for the standard 5-way 1-shot and 5-shot classification.   
Table~\ref{tab1: all experiments} compares the few-shot classification results between the first-order MAML and FORML with different backbone networks, against mini-ImageNet and CUB datasets. To have a fair comparison, FORML utilizes the same hyperparameter setting as MAML. Among all the experimental setups, FORML shows a significant improvement over its non-Riemannian counterpart (\textit{i.e} MAML) in terms of classification accuracy. Moreover, the presented results demonstrate the robustness of FORML against deeper backbones.

We also compare our method against its non-approximation-based version, RMAML \cite{tabealhojeh2023rmaml}, and two other types of methods including the approximation-based methods \cite{finn2017model,nichol2018first} and methods that focus on increasing the \textit{representation reuse} or \textit{representation change} \cite{raghu2019rapid,oh2021boil}. As this set of methods includes the Euclidean and Riemannian counterparts of FORML, it is important to compare them fairly. Thus, all experiments of FORML are conducted with the same hyperparameter settings as used for those methods. For Mini-ImageNet, Table~\ref{tab1: Mini_Conv4_experiments} demonstrates the effectiveness of FORML against the approximation methods (First-order MAML and Reptile). In addition, our proposed method shows superiority against ANIL and BOIL methods. Compared to its non-approximation-based version, RMAML, which is based on second-order gradients, FORML achieves a higher classification accuracy ($65.46\pm0.68$ vs $65.12\pm0.76$) in 5-shot classification, while obtains a competitive results in 1-shot classification.   
As presented in Table~\ref{tab:further-exp-CUB}, for the fine-grained dataset CUB, the proposed method outperforms all the considered methods on both 1-shot settings. In 5-shot experiments, FORML beats ANIL and BOIL methods and archives a competitive result, compared to RMAML.  

Adopting Conv-4 and ResNet-12 backbones, we also assess our algorithm against the FC100 benchmark. As shown in Table~\ref{tab:FC100-exp}, we observe that FORML improves the few-shot recognition performance on the FC100 dataset, compared to MAML and other methods. Surprisingly, our proposed method achieves competitive accuracy compared to other state-of-the-art methods on the FC100 dataset, as depicted in Table~\ref{tab:further-exp-FC100}.

Overall, single-domain experimental results show that using the same hyperparameters and setting, our proposed technique outperforms the original MAML and another approximation-based extension of MAML. Also, FORML demonstrates competitive results compared to its Riemannian non-approximation-based version, RMAML. Using the orthogonality constraints in the head of the models in FORML that limits the parameter search space, leads to higher performance and better convergence and acts as a regularization that prevents overfitting of the model. Moreover, learning the cosine similarity between the input and the parameters of the head facilitates the convergence and the adaptation of the head toward the optimum.    

\begin{table*}
\centering
\caption{Classification accuracies for \textbf{Omniglot$\rightarrow$EMNIST} and \textbf{mini-ImageNet$\rightarrow$CUB} cross-domain experimental setup.}\label{tab:cross-exp}
\begin{tabular}{llll}
\toprule
 & \multicolumn{2}{l}{\textbf{Omniglot$\rightarrow$EMNIST}} & \textbf{mini-ImageNet$\rightarrow$CUB} \\ 
 \cmidrule(r){2-4}
\textbf{Backbone Networks} & \multicolumn{2}{l}{\textbf{Conv-4}} & \multicolumn{1}{l}{\textbf{ResNet-18}} \\ 
\midrule

\textbf{Method} & \textbf{1-shot} & \textbf{5-shot} & \textbf{5-shot} \\ \cmidrule(r){1-1}
\cmidrule(r){2-3}
\cmidrule(r){4-4}
Baseline & $63.94\pm0.87$ & $86.00\pm0.59$ & \bm{$65.57\pm0.70$} \\
MAML & $72.04\pm 0.83$ & \bm{$88.24\pm0.56$}  & $51.34\pm0.72$ \\
FORML (ours) & \bm{$72.78\pm 0.82$} & $87.41 \pm0.57 $  &  $56.43\pm0.71$ \\ \bottomrule
\end{tabular}
\end{table*}

\subsubsection{Comparison with other state-of-the-art methods}
To further assess our proposed technique's performance, we compare FORML with additional few-shot learning algorithms. The results of mini-ImageNet, CUB, and FC100 datasets are depicted in Tables \ref{tab1: Mini_Conv4_experiments_sup}, \ref{tab:further-exp-CUB2}, and \ref{tab:further-exp-FC100}, respectively. Table \ref{tab1: Mini_Conv4_experiments_sup} shows the superiority of FORML performance compared to the state-of-the-art algorithms in 5-shot classification on the mini-ImageNet dataset. Moreover, our algorithm exhibits enhancements in 1-shot and 5-shot classification accuracy on the CUB dataset, compared to various state-of-the-art techniques utilizing ResNet-18 as the backbone architecture, as indicated in \ref{tab:further-exp-CUB2}. Furthermore, RMAML shows superior results in 5-shot classification on the FC100 dataset, outperforming all state-of-the-art algorithms.
\subsection{Cross-Domain Results}
We use the mini-ImageNet$\rightarrow$CUB and Omniglot$\rightarrow$EMNIST scenarios for conducting the cross-domain experiments. In the 1-shot Omniglot$\rightarrow$EMNIST experiment, FORML demonstrates superior performance compared to other approaches, as indicated in Table~\ref{tab:cross-exp}, while still achieving results similar to those of MAML for the 5-shot classification. Furthermore, FORML has significantly surpassed FOMAML by more than $5\%$ in terms of classification accuracy in the mini-ImageNet$\rightarrow$CUB experiment.
Nonetheless, as illustrated in Table~\ref{tab:cross-exp}, all methods are surpassed by the baseline. This occurs due to the inability of FORML's representation reuse to bridge the substantial domain gap between the meta-train and meta-test datasets, whereas the baseline technique (a transfer learning approach \cite{chen2019closer}) finds it much easier to adapt to the target domain by acquiring a new parameterized classifier layer ~\citep{vinyals2016matching, chen2019closer, tabealhojeh2023rmaml}. Despite this limitation, FORML continues to achieve competitive results in terms of classification accuracy when compared to other methods.

\subsection{Time and Memory Consumption}
To demonstrate the efficiency of our method, we also measured both time and video memory consumption during the meta-training, and compared them with their Euclidean (MAML) and non-approximation-based Riemannian (RMAML) counterparts, as depicted in Tables~\ref{tab1: time/memory} and \ref{tab1: time loops}. To do so, we utilized FORML (our proposed method), RMAML (the Riemannian non-approximation-based method), MAML (the Euclidean counterpart of RMAML), FOMAML (first-order approximation of MAML) to train a Conv-4 backbone on CUB dataset with the same hyperparameters such as number of inner-level steps, inner-level and outer-level learning rates and batch-size. The Results suggest that using FORML saves a tangible amount of time and memory as compared to its non-approximated version, RMAML. Moreover, although FORML uses Riemannian operations which are computationally expensive, it has a competitive memory usage and runtime as compared to Eucleadian methods. 

We also measured the time consumption of the baselearner and the meta-learner. The time consumption of the baselearner was measured over the forward process of the inner-loop, and the
time consumption of the meta-learner was measured over the backward processes. As illustrated in Table~\ref{tab1: time loops}, the execution time of one epoch of FORML is more than 3 times faster than RMAML. It also performs the inner-loop  20 times faster than RMAML. Compared with its Euclidean counterparts, FORML performed an epoch 12 times faster and the outer-loop updates 16 times faster than MAML, and achieved competitive runtime and memory usage compared to FOMAML.  
\begin{table}
\centering
\caption{Memory consumption}\label{tab1: time/memory}
\begin{tabular}{ll}
\hline
\multicolumn{1}{c}{\textbf{Method}} & \textbf{Memory (MB)} \\ \hline
MAML (first-order) &$1.81\times10^3$\\
MAML & $7.3\times10^3$\\
RMAML &$7.86\times10^3$ \\
FORML (ours)&$1.94\times10^3$\\
 \bottomrule
\end{tabular}
\end{table}

\begin{table*}[pt]
\centering
\caption{Total Execution time and average time of the inner and outer loops of 1 iteration (100 iterations)}\label{tab1: time loops}
\begin{tabular}{llll}
\hline
\multicolumn{1}{c}{\textbf{Method}} & \textbf{Execution time (seconds)}&\textbf{Avg. Inner-loop (seconds)} & \textbf{Avg. outer-loop (seconds)} \\ \hline
MAML (first-0order) &$7.62$& $6.09\times10^{-2}$ & $4.23\times10^{-3}$\\
MAML &$31.86$ &$6.19\times10^{-2}$ & $9.31\times10^{-1}$\\
RMAML &$39.80$& $9.50\times10^{-2}$ & $1.16$ \\
FORML (ours) &$12.14$& $9.03\times10^{-2}$ & $5.81\times10^{-2}$\\
 \bottomrule
\end{tabular}
\end{table*}
\section{Conclusion}
\label{Sec-Conclusion}
This paper introduced FORML, a first-order approximation method for meta-learning on Stiefel manifolds. FORML formulates a Hessian-free bi-level optimization, which reduces the optimization-based Riemannian meta-learning's computational cost and memory usage, and suppresses the overfitting phenomena. Using FORML to meta-learn an orthogonal fully-connected layer as the head of the backbone model, where its parameters are cast on the Stiefel manifold, has amplified the \textit{representation reuse} of the model during the meta-learning. By normalizing the input of the classification layer, a similarity vector between the input of the head and the class-specific parameters is calculated. As a result, the model learns how to move the head input toward the desired class. We conducted two experimental scenarios: single-domain and cross-domain few-shot classification. The results have demonstrated the superiority of FORML against its Euclidean counterpart, MAML, by a significant margin, while achieving competitive results against other meta-learning methods. We also compared FORML with its Riemannian and Euclidean counterparts in terms of GPU memory usage and execution time. Our proposed method demonstrates a significant improve over its non-approximation-based version, RMAML, by performing an epoch and an inner-level optimization loop more than 3 times and 20 times faster, respectievly. It also shows a remarkable improve over MAML and a competitive performance compared with FOMAML.

Motivated by the observed results, we aim to develop a multi-modal or multi-task  Riemannian meta-learning method in the future, that will be able to effectively tackle the cross-domain or even cross-modal scenarios.

\bibliographystyle{IEEEtran}
\bibliography{bibilography.bib}

\begin{IEEEbiography}{Hadi Tabealhojeh}{\space} is a Ph.D. candidate in Artificial Intelligence and Robotics working under the supervision of Dr. Peyman Adibi and Dr. Hossein Karshenas in the Artificial Intelligence Department, Faculty of Computer Engineering, University of Isfahan, Isfahan, Iran. He received his M.Sc. degree from Shahid Chamran University of Ahvaz, Ahvaz, Iran. He is currently a member of the Intelligent and Learning Systems Research Lab. His current research interests include Deep Learning, Meta-Learning, and multi-modal learning, especially on curved and non-linear spaces and their applications. Currently, He is a lecturer at the Department of Computer Engineering, Faculty of Engineering, Shahid Chamran University of Ahvaz.
\end{IEEEbiography}
\begin{IEEEbiography}{Soumava Kumar Roy}{\space} received his Ph.D. degree from the College of Engineering and Computer Science, Australian National University in 2021. Before this, he received his bachelor’s of engineering degree in electronics and communication engineering from the Manipal Institute of Technology, Manipal, India, in 2013, and a master’s of technology degree in Information Technology from the Indian Institute of Information Technology, Allahabad, India. His research interests include deep learning and computer vision. He is currently working as a post-doctoral researcher in the Computer Vision Lab, EPFL. 
\end{IEEEbiography}
\begin{IEEEbiography}{Peyman Adibi}{\space} received the Ph.D. degree from the Faculty of Computer Engineering, Amirkabir University of Technology, Tehran, Iran. He is currently an Associate Professor at the Artificial Intelligence Department, Faculty of Computer Engineering, University of Isfahan, where he is the head of the Intelligent and Learning Systems Research Lab. He was a visiting Professor at GIPSA Lab, Grenoble Institute of Technology, Grenoble, France, from 2016 to 2017. His current research interests include Machine Learning and Pattern Recognition, Multimodal and Geometric Learning, Computer Vision and Image Processing, Computational Intelligence and Soft Computing, and their applications.
\end{IEEEbiography}
\begin{IEEEbiography}{Hossein Karshenas}{\space} received his BE degree in computer engineering from Shahid Beheshti University, Tehran, Iran, in 2006 and ME degree in artificial intelligence and robotics from Iran University of Science and Technology (IUST), Tehran, Iran, in 2009. He received his PhD degree in artificial intelligence from the Technical University of Madrid (UPM), Madrid, Spain, in 2013. He is currently an assistant professor at the Artificial Intelligence Department, Faculty of Computer Engineering, University of Isfahan, Isfahan, Iran. His main research interests include estimation of distribution algorithms, computational intelligence, data analytics and modeling, and multi-objective optimization where he has published several articles in peer-reviewed journals.
\end{IEEEbiography}

\end{document}